%% file: naacl2021.tex
\pdfoutput=1
\documentclass[11pt]{article}

\usepackage{naacl2021}

\usepackage{times}
\usepackage{latexsym}

\usepackage[T1]{fontenc}
\usepackage[utf8]{inputenc}

\usepackage{microtype}

\usepackage{graphicx}
\usepackage{amsmath}
\usepackage{footmisc}
\usepackage{multirow}

\usepackage{ulem}
\usepackage{float}
\restylefloat{table}

\title{Distantly Supervised Transformers For E-Commerce Product QA}

\author{Happy Mittal \\
  India Machine Learning \\
  Amazon \\
  \texttt{mithappy@amazon.com} \\\And
  Aniket Chakrabarti \\
  India Machine Learning \\
  Amazon \\
  \texttt{chakanik@amazon.com} \\
  \And
  Belhassen Bayar\Thanks{This work was done while author was in Community Shopping team.} \\
  Digital Video \\
  Amazon \\
  \texttt{bayarb@amazon.com} \\
  \AND
  Animesh Anant Sharma\\
  Community Shopping\\
  Amazon \\
  \texttt{shanimes@amazon.com}\\
  \And
  Nikhil Rasiwasia\\
  India Machine Learning \\
  Amazon \\
  \texttt{rasiwasi@amazon.com} \\
}

\begin{document}
\maketitle

\input{abstract}
\input{intro}

\input{related}

\input{semantic}
\input{expt}

\input{conclusion}

\input{ack}
% \clearpage
\bibliography{anthology,references}
\bibliographystyle{acl_natbib}

%\appendix

%\section{Example Appendix}
%\label{sec:appendix}

%This is an appendix.

\end{document}

%% file: abstract.tex
\begin{abstract}
We propose a practical instant question answering (QA) system on product pages 
of e-commerce services, where for each user query, relevant community question answer (CQA) 
pairs are retrieved. User queries and CQA pairs differ significantly in language
characteristics making relevance learning difficult. Our proposed transformer-based model learns a
robust relevance function by jointly learning unified syntactic and semantic 
representations without the need for human labeled data. This is achieved by 
distantly supervising our model by distilling from predictions of a syntactic 
matching system on user queries and simultaneously training with CQA pairs. 
Training with CQA pairs helps our model learning semantic QA relevance and distant 
supervision enables learning of syntactic features as well as the nuances of user 
querying language. Additionally, our model encodes queries and candidate 
responses independently allowing offline candidate embedding generation thereby 
minimizing the need for real-time transformer model execution. Consequently, our
framework is able to scale to large e-commerce QA traffic. Extensive evaluation 
on user queries shows that our framework significantly outperforms both 
syntactic and semantic baselines in offline as well as large scale online A/B 
setups of a popular e-commerce service.
\end{abstract}

%% file: intro.tex
\section{Introduction}
\label{sec:intro}
\begin{figure}[h]
  \begin{center}
    \includegraphics[scale=0.6]{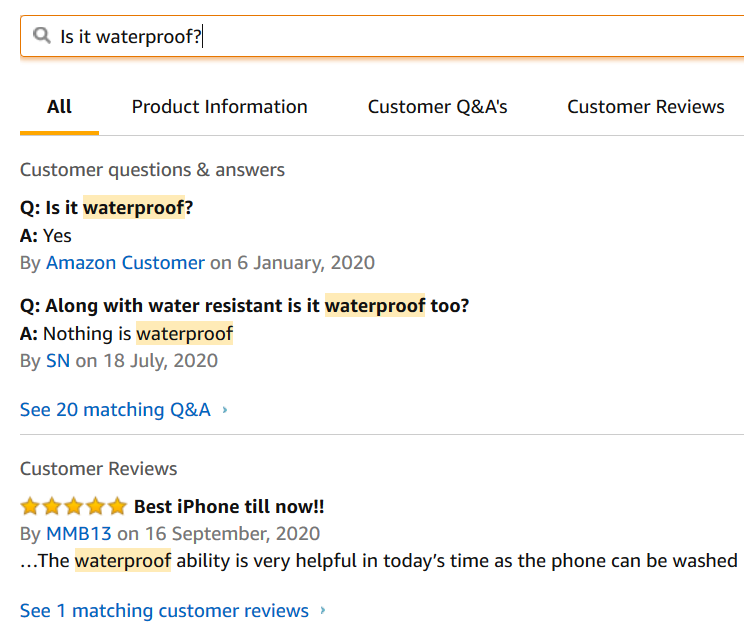}
  \end{center}
  \caption{Instant QA widget on Amazon}
  \label{fig:amazon-swdp}
\end{figure}

Product pages on an e-commerce service (eg. Amazon) are often overloaded with 
information. 
Customers wanting to search for a piece of specific information about a product find it 
difficult to sift through. 
To address this issue most services provide an instant QA system on the product
pages 
enabling users to type their query and get instant answers 
curated from various sources present on the page. 
Figure~\ref{fig:amazon-swdp} shows the QA widget on Amazon, and
the three sources viz. Product information (eg: bullet points, 
technical specifications etc.), Customer Q\&A's (where customers/sellers provide 
an answer to the posted questions by customers, henceforth called community QA
or CQA section), and Customer reviews from
where a response is generated. 
In this paper, we focus on retrieving responses
from the CQA section. Hence our goal is to learn a robust relevance
function between user queries and CQA pairs. Notably, these two domains differ
significantly in language characteristics. User 
queries are typically short, often ill-formed and incomplete, whereas
CQA pairs tend to be more complete and well-formed. For example,
"Bettry perfon" is a user query where the intended question probably was "how is
the battery performance?". Furthermore, we analyzed CQA section along with 3 months
user query logs of a popular e-commerce service and found that the data 
statistics such as length, vocabulary overlap (between user queries and CQA)
indicate that the domains are quite different.
%ssaw that IN locale of
%a popular e-commerce platform we analyzed the CQA section of all products and
%3 months of user queries and found that the average user query length is 3.06
%words as compared to 8.56 for questions in the CQA section. Even the user query
%and questions posted in CQA section have a vocabulary overlap of less than 40\%. 
Consequently, relevance
learning for this task is difficult. Table~\ref{table:lang-characteristics}
characterizes these differences %chaknik::\textcolor{red}{on 3 months of user query data (from search logs)} 
for 4 different locales: Canada (CA), 
Germany (DE), France (FR), and IN (India).
% , and table~\ref{table:in-top-words}
% contains the top frequently used words in user queries and CQA pairs (excluding
% stop words) for IN locale.
\begin{table}[h]
\begin{tabular}{|l|c|cc|}
\hline
   & \multicolumn{1}{c|}{\multirow{2}{*}{\begin{tabular}[c]{@{}c@{}}Vocab Overlap\\Percentage\end{tabular}}} & \multicolumn{2}{c|}{\begin{tabular}[c]{@{}c@{}}Avg. Length\end{tabular}} \\ \cline{3-4}
   \multicolumn{1}{|c|}{} & \multicolumn{1}{c|}{}   & User Query & \begin{tabular}[c]{@{}c@{}}CQA\\Question\end{tabular}      \\ \hline
CA & 55.4   & 3.26  & 11.79    \\
DE & 59.4   & 2.58  & 12.05   \\
FR & 59.1   & 4.21  & 13.79    \\
IN & 39.6   & 3.06  & 8.56    \\ \hline
\end{tabular}
\caption{Differences in user queries and CQA}
\label{table:lang-characteristics}
\end{table}

% \begin{table}[h]
% \begin{tabular}{|l|}
% \hline
% Common user query words                       \\ \hline
% word1, word2 \\ \hline
% Common CQA words                       \\ \hline
% word1, word2                       \\ \hline
% \end{tabular}
% \caption{Frequently used words in user queries and CQA in IN locale}
% \label{table:in-top-words}
% \end{table}
Existing QA systems typically work by retrieving a set of candidates for a user
query using syntactic features (eg. BM25 that uses bag of words features) followed by a semantic answer 
selection/re-ranking step~\cite{Chen2017ReadingWT}. Some approaches include 
semantic features in the candidate generation step~\cite{Mitra2019AnUD}.
Syntactic systems fail in two cases: (1) when there are no word overlaps (a
likely scenario as user queries have limited vocabulary overlap with CQA pairs), 
and (2) when the word overlaps are semantically irrelevant. 
While adding semantic features or semantic re-ranking models mitigate some
of the drawbacks, however, training a robust semantic relevance model to match
user queries with CQA pairs is difficult due to the lack of human-labeled data.
An additional challenge is that the instant QA system needs to
provide real-time responses to users and must scale to the very large traffic
of modern e-commerce systems. Running deep models online (typical in case of
re-ranking) is prohibitive for
such a system.
 
In this paper, we present an instant QA system with two main contributions: (1) 
our framework is able to learn a robust relevance function between user queries
and CQA pairs by jointly learning semantic and syntactic features-aware representations
without the need for explicit human-labeled data, and (2) our framework 
minimizes the need for real-time model execution by encoding the CQA pairs offline,
enabling large scale online deployment.

We chose BERT~\cite{Devlin2019BERTPO} as our 
transformer encoder due to
its recent success in various natural language understanding (NLU) 
tasks including QA. 
To address the lack of labeled training data challenge, we use
the QA pairs from the CQA section of each product page as training data.
However, as shown in our evaluation (section~\ref{subsec:offline-eval}), 
such a model does not work well on the
user queries asked on the instant QA system on the product pages. 
We propose a distillation-based distantly supervised training algorithm
where we use the answers retrieved by a syntactic match system on a set of
user queries asked on the instant QA system. This training helps the model
adapt to the specific task at hand by learning the user query distribution as
well as the strengths of a traditional syntactic match system. This coupled
with training on CQA pairs helps our model learn a
robust semantic model that is task aware. Our training data does not require
any explicit human labeling.

To make our system work in real-time we train the BERT model in Siamese 
style~\cite{Reimers2019SentenceBERTSE} with triplets consisting of query, relevant
candidate (+ve sample), and irrelevant candidate (-ve sample).
Hence the query and candidate responses are encoded independently using the same
transformer encoder enabling embedding computation of all candidates (across all
products) offline. At real-time, only the user query needs to be embedded using
the heavy semantic model resulting in a significant reduction of online compute
cost.  In contrast, the common practice of using BERT in QA problems is to
concatenate the query and a candidate response and run BERT on the fused input.
This would require BERT to run on all query, candidate CQA pairs on product
pages real-time making it prohibitive for online deployment.
Additionally, we combine the two embeddings (question and 
answer) in each CQA pair to
form one embedding per pair allowing us to reduce the offline storage 
significantly.

We extensively evaluate our framework on user queries asked on the instant QA 
system at a popular e-commerce system 
in 4 locales spanning 3 languages.
Offline evaluation  
shows that our proposed framework is able to increase the area under the
precision-recall curve (PR-AUC)
by up to 12.15\% over the existing system. 
Also in an online A/B test, our system is able to improve
coverage by up to 6.92\% by complementing
the existing system.

%% file: related.tex
\section{Related Works}
\label{sec:related}

\noindent \textbf{QA Systems:}
Question Answering (QA) is a fundamental task in  
the Natural 
Language Understanding (NLU) domain. Broadly QA systems can be categorized into
open-domain QA and closed-domain QA. Open-domain QA involves answering 
questions related to all topics from a huge repository of information such as
the Web~\cite{Voorhees1999TheTQ},
Wikipedia corpus~\cite{Yang2015WikiQAAC},  Knowledge Bases
~\cite{Bollacker2008FreebaseAC}. 
Closed-domain QA systems usually deal with
a specific domain such as medical, sciences etc. The main steps of a QA
system are candidate retrieval followed by answer selection/re-ranking
~\cite{Chen2017ReadingWT}. Some systems do answer 
generation~\cite{Lewis2020BARTDS} instead of
selection.
%followed by span
%highlighting (MRC~\cite{Rajpurkar2016SQuAD10}) and 
%(2) natural language answer 
%generation~\cite{Lewis2020BARTDS,Raffel2019ExploringTL}.

\noindent \textbf{Semantic Text Encoders:}
Recently, QA systems have significantly evolved from syntax based (eg. BM25)
systems to leverage the power of semantic text representation models. Recurrent
Neural Networks (RNN) such as Long Short Term Memory 
(LSTM)~\cite{Hochreiter1997LongSM} networks were defacto for semantic
text representation. Recently proposed self attention based 
transformer~\cite{Vaswani2017AttentionIA} models
show consistent improvement over RNNs on a multitude of NLU tasks such as
Machine Translation (MT)~\cite{Vaswani2017AttentionIA}, Machine Reading
Comprehension~\cite{Rajpurkar2016SQuAD10},
GLUE~\cite{Devlin2019BERTPO} and Natural Language Generation (NLG) 
tasks~\cite{Radford2019LanguageMA}. %The span finding task of question answering
%saw significant improvements by leveraging the transformer 
%architecture~\cite{Devlin2019BERTPO,Liu2019RoBERTaAR,Yang2019XLNetGA}.

\noindent \textbf{E-commerce Product QA Systems:}
E-commerce Product QA systems are similar
to domain specific systems. Recently product QA systems are receiving a lot of 
attention
due to their growing usage and unique characteristics such as the search space
being specific to each product.
Product QA systems are real-time systems where
a user types a query and expect instant answers, the
queries of such systems are typically short, prone to errors and even incomplete
in nature. This coupled with product specific limited search space, often
results in no syntactic match between
the query and candidate answers, making semantic matching essential. In contrast, the 
retrieval set for websearch and traditional IR typically is huge and there are 
always bag-of-words matches that are used to filter down the candidates before 
running subsequent deep models. Additionally, search and IR systems in 
e-commerce/web domains get 
powerful implicit supervision signals through user clicks, however, instant QA 
on product pages only show the answer with no option to click making it hard to 
get user feedback based labels. Finally, QA relevance is different from 
traditional IR relevance (eg. for query “what is the material?”, the response 
“made of stainless steel” 
is relevant and doesn’t require bag-of-words or even synonym matches) making 
domain specific semantic matching critical.
Kulkarni
et al.~\cite{kulkarni2019productqna} propose an embedding
based semantic matching model to find relevant answers. 
Additionally, it uses a query category classifier
and an external ontology graph both of which require human generated labels.
There are several proposed 
works~\cite{Zhang2020AnswerRF,Zhang2019DiscoveringRR,Zhang2020LessIM,Chen2019AnswerIF,McAuley2016AddressingCA,Burke1997QuestionAF,Gupta2019AmazonQAAR}
that improve the
QA relevance models (usually learned from CQA pairs) by enriching them using 
information from reviews
of the product and capturing their relation with the CQA pairs.
Natural language answer generation models are also used in the context of
product 
QA~\cite{Deng2020OpinionawareAG,Chen2019ReviewDrivenAG,Gao2019ProductAwareAG,Bi2019IncorporatingEK}. 
They are typically encoder-decoder 
architectures and their variants. These models are hard to 
generalize and often
result in factually incorrect text generation. The aforementioned works use
reviews and other product information along with CQA section
to guide the models to generate answers.
%\textcolor{red}
{%Note that the product QA systems differ from the traditional search/information retrieval in that search in e-commerce/web domains get 
%powerful supervision signals through clicks, however, instant QA on product pages only show the answer with no option to click making it hard to 
%get user feedback based labels. Consequently learning a domain specific product QA relevance function is hard. Existing semantic matching 
%~\cite{Reimers2019SentenceBERTSE,Reimers2020MakingMS} strategies don't generalize due to the following unique nuances: (1) user querying language 
%is significantly different from CQA (eg. low vocabulary overlap between user queries and CQA, table~\ref{table:lang-characteristics}), and (2) 
%QA relevance is different from traditional IR relevance (eg. for query “what is the material?”, response like “made of stainless steel” 
%is relevant and doesn’t require bag-of-words or even synonym matches) making domain specific semantic matching critical. 
%In contrast the 
%retrieval set for websearch typically is huge and there are always bag-of-words matches that are used to filter down the candidates before 
%running subsequent deep models. 
%Additionally, user queries/CQA characteristics of different product categories are also very different and 
%we address the challenging problem of building a single model that generalizes to all products.}

In this paper, we take the approach of answer retrieval (instead of 
generation). We 
%propose to 
solve the orthogonal problem of how to adapt
the relevance model to be aware of the user query characteristics 
(significantly different from the well formed questions posted in the 
CQA section) in the absence of human labeled data.
The improvement in relevance 
models (between user 
queries and CQA pairs) proposed 
%here 
can be easily complemented with
the existing review awareness models. A drawback of the aforementioned models is
they comprise of multiple deep neural components, many of which need to be
run real-time making
online model deployment and computation cost prohibitive for large scale
deployment. Our 
%proposed 
framework
only needs to encode the user query realtime, all candidate
responses are pre-computed stored in an index making it amenable to real-time
deployment.

%% file: semantic.tex
\section{Semantic QA System}
\label{sec:semqa-system}
In this section, we describe our proposed semantic QA system for e-commerce
services. Unlike traditional QA systems where multiple models are used
sequentially to surface the final response (eg. candidate retrieval, followed
by answer selection/re-ranking, followed by span selection),
here we use a
semantic index and the top results retrieved from the index are the final
answers shown to the users. Below we describe the problem definition followed by
individual components of our
system:

\subsection{Problem Statement}
 Given a set of $N$ products, a user query $u_q$ on product $p$ and 
 the set of CQA pairs for all products 
 $C = \{\{Q,A\}_p\}$ where $p \in \{1,N\}$ and $\{Q, A\}_p = 
 \{\{q, a\}_p^1, \{q, a\}_p^2, ... \{q, a\}_p^n\}$
 are the set of $n$ QA pairs for product $p$, the goal is to find the relevant
 QA pairs set $R \subseteq C$ such that $\forall \{q,a\} \in C, \{q,a\}$  
 can answer $u_q$.

\subsection{Model Architecture}
\label{subsec:textrep}
We chose the transformer network~\cite{Vaswani2017AttentionIA} as our core text 
representation model. Transformers are largely successful in QA systems (eg.
BERT for MRC~\cite{Devlin2019BERTPO}), however, the typical approach to use
transformers in a QA setting is to create a single input concatenating both the 
user query
and a candidate response, enabling transformers to leverage a full
contextual representation through attention mechanisms. Since transformer models
are usually very large (hundreds of millions of parameters), this makes it infeasible
to run the model real-time on a large candidate set. Our goal in this work is to leverage
the strengths of the deep representational power of transformers while being
able to scale to a real-time system with large candidate sets.
%During to this real-time
%
%Transformer based language models have been known to
%perform well in the MRC based QA
%systems~\cite{Devlin2019BERTPO,Liu2019RoBERTaAR,Yang2019XLNetGA}. However, these
%models require the user query and candidate response concatenated as a single 
%input. 
%Since the user query is unknown beforehand, the model needs to be
%executed real-time for a user query with all the 
%candidate
%responses for that product. 
%Transformer
%based language models usually have hundreds of millions
%of parameters and computing them real-time for all question-candidate combination
%is infeasible in terms of latency and cost.
%Consequently, most systems decouple the candidate
%retrieval step and the answer selection/re-ranking step.
Hence we propose to
use transformers in a Siamese network 
setting similar to Sentence BERT~\cite{Reimers2019SentenceBERTSE} to embed the 
query and the
candidate responses independently. The same transformer encoder is used to encode
both the query as well as the candidate responses (CQA pairs). This
enables offline encoding of all CQA pairs and at real-time only, the user 
query needs to be encoded making the model productionizable at scale.

In our model, a sequence of text is encoded first by passing it through the
transformer network that generates embeddings for each token in the input
sequence. 
A mean pool (average) of the output token embeddings 
% is used to represent 
represents
the
full sequence.
%Let $t_1,t_2,....t_k$ be the sequence of input tokens, $$
\begin{equation}
\label{eq:emb_eqn}
e(text) = meanpool(transformer(text))
\end{equation}
We train our transformer based QA system using the triplet 
loss~\cite{Chechik2009LargeSO} that tries to learn
a similarity measure between user query, CQA pairs while maximizing the margin 
between relevant pairs and irrelevant pairs. Such ranking loss 
 has proven effective at numerous ranking 
 tasks~\cite{Chechik2009LargeSO,Schroff2015facenet,Wang2014ranking}. 
 The triplet loss for a %\sout{anchor}\textcolor{red}{user} 
 query $q$ (also known as anchor), a relevant candidate response
 $c_{+ve}$, and an irrelevant candidate response $c_{-ve}$, is formally 
 defined as:
 \begin{align}
 \label{eq:loss}
 \max\big(||e(q) - &e(c_{+ve})|| - \nonumber \\
 &||e(q) - e(c_{-ve})|| + 
 \epsilon, 0\big)
 \end{align}
 where $||\cdot||$ is the Euclidean distance, and $\epsilon$ is the 
 margin. The goal is to maximize the loss over the triplets of the training
 set.
 
\subsection{Distantly Supervised Training}
\label{subsec:ds-training}
One of the biggest challenges in training the instant QA system for an 
e-commerce service is the lack of task specific labeled data. One source of 
labeled data is
the CQA pairs. To create the relevant pairs (positive samples) and
irrelevant pairs (negative samples) we adopt the following sampling strategy:
(1) we sample 
% \sout{anchor}
user questions (as anchors)
from all product pages' CQA section. This
ensures the diversity of products in the training data. (2)
For each question, we pick a paired answer to that question as the
relevant pair. (3) For the same 
% \sout{anchor}
user question, we randomly
select negative samples (answers from different user questions) both from the same product page and from other product
pages. The negatives from the same product page are the hard negatives %i.e. they have high probability of syntactic match with
%the positive answer, hence harder to distinguish
(as these answers are related to the current product whereas answers from other
product pages likely are completely unrelated and easy to distinguish). In future,
 we wish to explore advanced negative sampling strategies such as Kumar et al.~\cite{Kumar2019ImprovingAS}
 for answer sampling. However,
for pages having very few CQA pairs, the number of negative samples
becomes small, and adding negative samples from other product pages is useful in
such scenarios even though those may be easy negatives. We show (in 
section~\ref{subsec:offline-eval}) that such a 
model learns a good QA relevance function (between community questions and
answers), 
however, it fails to learn a robust relevance function between
the typical user queries asked on the instant QA widget and the CQA
pairs (candidate responses). The underlying reason is the difference in 
characteristics
of the questions/answers posted in CQA forum (typically long, 
well-formed, and
complete) and the queries asked on the instant answer widget (often short,
grammatically incorrect, and ill-formed). 
Consequently, a model trained to
learn relevance between community questions and answers performs very well when
the queries are long and well-formed, however, they perform poorly on the 
queries typically asked by a user on the instant answer widget.

\vspace{-0.01cm}
To address the aforementioned challenge, we propose a knowledge 
distillation~\cite{Hinton2015DistillingTK} based training technique that
acts as distant supervision on our Siamese transformer network.
We collect a random set of user queries
asked on the instant QA system 
and the responses (CQA pairs) generated by the existing syntactic match
system from the query logs of a popular e-commerce service.
For generating the relevant pairs we take a user query as the
anchor question and the answer from the CQA pair retrieved
by the existing system. For generating the irrelevant pairs we follow a similar
negative strategy as before. %of sampling negatives from the same product and from across 
%products.
The existing syntactic match based system can be thought
of as the teacher model and the Siamese transformer model is the student model
in the distillation process. This distant supervision helps our semantic model
adapt to the nuances of the instant QA system where queries are often
short, and incoherent. Additionally, the distant
supervision system also helps the semantic model learn the strengths of 
syntactic match systems. 

We train our Siamese transformer network with data from both the aforementioned
sources (CQA pairs, distilling from predictions of syntactic match
based system on real user queries). We explore two strategies for jointly training 
our model with the two data sources: (1) we mix the data from both sources and
train our model with the single triplet loss, and (2) we train our model in a 
multi-task fashion where there is a task (triplet loss) for each of the two
data sources. This joint training of a unified syntactic and semantic 
representation
while adapting to the nuances of user querying language enables our instant QA 
system to learn a robust task specific relevance function.
% Consequently
Hence
our instant QA system serves as an end-to-end
unified framework for the e-commerce product QA problem.

\subsection{Model Inference}
For our proposed model the input is a user query on the instant QA system.
The query is embedded in real-time using equation~\ref{eq:emb_eqn} and
searched against the candidate vectors (for that specific product) to retrieve 
the top-k most relevant candidates (where a candidate is an embedding of QA pair 
from the CQA section of the product). For the top-k search, we use a
weighted combination of squared Euclidean distance between the query, question 
(of CQA pair) 
embeddings and query, answer (of CQA pair) embeddings. Our relevance score
of a query, CQA pair is generated as follows:
\begin{align}
\label{eqn:rel_score}
s(q,Q,A) = \alpha &||e(q) - e(Q)||^2 \nonumber\\
&+(1-\alpha) ||e(q) - e(A)||^2 
\end{align}
The above expression can be
rewritten using linearity of inner products as follows:
%\begin{equation}
\begin{align}
\label{eqn:rel_score_simple}
%s(q,Q,A) &= %\alpha ||emb(q) - emb(Q)||^2 + 
%(1-\alpha) ||emb(q) - emb(A)||^2 \nonumber \\
%&= \alpha||emb(q)||^2 + \alpha||emb(Q)||^2 - 2\alpha \langle emq(q),emb(Q) \rangle \nonumber \\
%&\ \ +(1-\alpha)||emb(q)||^2 + (1-\alpha)||emb(A)||^2 -2(1-\alpha) \langle emb(q),emb(A) \rangle \nonumber \\
||e(q)||^2&+\alpha||e(Q)||^2+(1-\alpha)||e(A)||^2 - \nonumber \\
&\qquad 2 \langle e(q),\alpha e(Q) + (1-\alpha) e(A) \rangle
\end{align}
Here $\langle \cdot , \cdot \rangle$ denotes the inner product between vectors.
From the expression in equation~\ref{eqn:rel_score_simple} we can see that
instead of storing $e(Q)$, and $e(A)$ separately, we can store the
weighted combination of the two vectors $\alpha e(Q) + (1-\alpha) e(A)$
along with two extra scalar dimensions $\alpha||e(Q)||^2$ and
$(1-\alpha) ||e(A)||^2$ and the rest of the terms are query related and are 
computed
real-time. This enables us to reduce the offline index storage by half by storing
only one vector per candidate QA pair. Note that to enable such relevance score
computation we had to use the square of Euclidean distance (instead of vanilla
Euclidean distance) as the relevance
scoring function at inference time.

%% file: expt.tex
\section{Experiments}
We ran experiments both in offline settings as well as in
large scale online setups. We evaluated our models across 4 
locales with 3 languages to test 
whether our distant supervision
based training approach is able to generalize across languages and varying data 
characteristics.

\subsection{Methods}
In this section, we describe the methods that we compare. All
methods described below can encode query and candidates independently.
Consequently, the candidate index may be computed offline for all of these
methods, enabling large scale deployment.

\noindent \textbf{BM25:} BM25~\cite{Robertson1994OkapiAT} is the defacto ranking
	function used in retrieval systems. It relies on a weighted combination
	of Term Frequency (TF) and Inverted Document Frequency (IDF) matching. The
	standard form of the scoring function is as follows:
	% \begin{align*}
	% &bm25(q,D) = \\
	% &\qquad\sum_{i=1}^{n} IDF(q_i) \frac {TF(q_i, D) (\gamma + 1)} {TF(q_i, D) + \gamma (1 - \delta + \delta \frac{|D|}{\overline{|D|}}) } \\
	% &where,\ IDF(q_i) = ln(\frac{N-|D(q_i)|+0.5}{|D(q_i)|+0.5}+1)
	% \end{align*}
	% Here $q$ is the customer query consisting of $n$ terms ($q_1, q_2,\dots 
	% q_n$), $D$ is document (or a sequence of text) consisting of $k$ terms,
	% $|D|$ denotes the number of terms in document D, $\overline{|D|}$ is the 
	% average
	% number of terms per document, $|D(q_i)|$ is the number of documents
	% containing the term $q_i$, $N$ is the total number of documents in the
	% corpus, and $\gamma$, $\delta$ are tunable parameters. 
	\begin{align*}
	&bm25(q,D) = \\
	&\qquad\sum_{i=1}^{n} IDF(q_i) \frac {TF(q_i, D) (k + 1)} {TF(q_i, D) + k \left(1 - b + b \frac{|D|}{avgdl}\right)} \\
	&where,\ IDF(q_i) = ln\left(\frac{N-m(q_i)+0.5}{m(q_i)+0.5}+1\right)
	\end{align*}
	Here $q$ is the user query consisting of $n$ terms ($q_1, q_2,\dots 
	q_n$), $D$ is a document (or a sequence of text), $TF(q_i,D)$ denotes the number of times $q_i$ appears in $D$,
	$|D|$ denotes the number of terms in document D, $avgdl$ is the 
	average number of terms per document, $m(q_i)$ is the number of documents
	containing the term $q_i$, $N$ is the total number of documents in the
	corpus, and $k$, $b$ are tunable parameters, which we fixed to 1.5 and 0.75 
	respectively~\cite{manning2008introduction}.
	Given the $bm25$
	function above, we derive the relevance function between a user query,
	and a CQA pair in a similar fashion as equation~\ref{eqn:rel_score}
	as follows:
	\[
	\alpha bm25(q,Q) + (1-\alpha) bm25(q,A) 
	\]

\noindent \textbf{E-commerce Baseline:} We use the syntactic feature 
based
existing optimized instant QA system at a popular e-commerce service as a baseline. 
We collect the query and 
responses shown by the system from query logs.

\noindent \textbf{Sentence-transformers-STS-NLI}: We use 
	sentence-transformers~\cite{Reimers2019SentenceBERTSE,Reimers2020MakingMS}
	which
	are state-of-the-art Siamese style trained transformer models  
	for the general 
	purpose semantic textual similarity (STS) and natural language inference 
	(NLI) task. For English, we use the
	roberta-large-nli-stsb-mean-tokens
	\footnote{\label{fn:sbert-url}https://www.sbert.net/docs/pretrained\_models.html} 
	%\footnote{https://public.ukp.informatik.tu-darmstadt.de/reimers/sentence-transformers/v0.2/roberta-large-nli-stsb-mean-tokens.zip} 
	model, and for French and German 
	we use the xlm-r-100langs-bert-base-nli-stsb-mean-tokens 
	\footref{fn:sbert-url} model as we found
	them to be the best performing pretrained models. The relevance function is
	computed in a similar fashion as equation~\ref{eqn:rel_score}.

\noindent \textbf{SemQA-CQA:} Our proposed model trained
	only with CQA data as described in section~\ref{sec:semqa-system}.

\noindent \textbf{SemQA-CQA-DS:} Our proposed model that was trained
	with CQA data and distantly supervised with predictions of
	syntactic match system on user queries as described in
	section~\ref{sec:semqa-system}.
	
\subsection{Training Setup}
We collect training data from the CQA section and user query logs
for CA, DE, FR and IN locales of a popular e-commerce service. 
For each locale, to generate the CQA triplets and user query triplets (for 
distant supervision), we use data from CQA section of products, and user query
logs and follow the sampling strategy described in 
section~\ref{subsec:ds-training}.
%CQA triplets (community question, positive answer, negative answer), we first
%collect the community QA pairs from product pages of the respective locale and
%then apply the positive and negative sampling technique described in
%section~\ref{subsec:ds-training}. To generate the training data for 
%distant supervision, for
%each locale we collect the user queries and the relevant CQA pair
%predicted by the existing syntactic match based system. Again we use the 
%positive and negative
%sampling strategy as described before to create the triplets for distant
%upervision. 
The dataset statistics are described in table~\ref{table:train-data-stats}.
\begin{table}[h]
\begin{tabular}{|l|cc|}
\hline
   & \multicolumn{1}{c}{\begin{tabular}[c]{@{}c@{}}CQA Triplets\end{tabular}} & \multicolumn{1}{c|}{\begin{tabular}[c]{@{}c@{}}User Query Triplets\end{tabular}} \\ \hline
CA & 5,317,904   & 1,063,580      \\
DE & 5,000,000   & 4,949,766      \\
FR &  1,500,000  &   173,258    \\
IN &  7,176,824  &  10,641,498     \\ \hline
\end{tabular}
\caption{Training data statistics}
\label{table:train-data-stats}
\end{table}

We use the bert-base-uncased\footnote{https://huggingface.co/bert-base-uncased} 
as the base transformer for our English models
(for CA and IN locale), camembert-base~\cite{Martin2020CamemBERTAT}
\footnote{https://huggingface.co/camembert-base} as the base transformer for FR 
locale,
and bert-base-multilingual-uncased
\footnote{https://huggingface.co/bert-base-multilingual-uncased} 
as the base transformer for DE locale. We
train our models upto 10 epochs, with a batch size of 16, Adam
optimizer with learning rate
of $2e-5$ with a schedule of linear warmup of first 10000 steps and then
linear decay. We set $\epsilon=1$ in the loss equation~\ref{eq:loss}, and $\alpha=0.4$
in the inference equation~\ref{eqn:rel_score}. For the joint training (CQA 
triplets and user query triplets), we have two training runs (data mixing and multi-task
as described in section~\ref{subsec:ds-training}) per locale and picked
the best models (data mixing for CA, FR and multi-task for DE, IN). 
We use the Pytorch\footnote{https://pytorch.org}, 
Huggingface~\cite{Wolf2019HuggingFacesTS} and
Sentence-Transformers~\cite{Reimers2019SentenceBERTSE} libraries to develop our
models on an Nvidia
V100 GPU and hence our training time per batch and inference time per sample
are same as that of Sentence-Transformers with BERT (base-model, 110M
parameters).
%\end{itemize}

\subsection{Offline Evaluation}
\label{subsec:offline-eval}
We do offline evaluation of our models under two settings: (1) on CQA
test sets collected from the product pages at a popular e-commerce service% where the queries and responses
%are well formed, complete, and
%semantic intent is easily understable
, and 
(2) on user queries test set collected from query logs of the instant QA
system on product
pages of the same e-commerce service.% where queries are typically short with grammatical errors and often
%incomplete.
Table~\ref{table:test-data-stats} contains the test data statistics.
\begin{table}[h]
\begin{tabular}{|l|c|cc|}
\hline
   & \multicolumn{1}{c|}{\begin{tabular}[c]{@{}c@{}}CQA \\Test Set\end{tabular}} & \multicolumn{2}{c|}{\begin{tabular}[c]{@{}c@{}}User Queries\\Test Set\end{tabular}} \\ \hline
   & \#Questions   & \#Queries & \begin{tabular}[c]{@{}c@{}}\#Query-Response \\Pairs\end{tabular}      \\ \hline
CA & 2722   & 1485  & 5992    \\
DE & 2871   & 1351  & 5591   \\
FR & 2547   & 1762  & 5127    \\
IN & 2773   & 1459  & 4225    \\ \hline
\end{tabular}
\caption{Test data statistics}
\label{table:test-data-stats}
\end{table}

\noindent \textbf{Evaluation on CQA Dataset:}
The goal of this section is to evaluate the relevance between community 
questions and answers learned by
different approaches. For all locales
we randomly sample questions posted on product pages. The paired answers to
those questions are considered to be relevant answers and all other answers
(from other CQA pairs) of the product are assumed to be irrelevant answers. We
only sampled products that at least have 5 CQA pairs posted. For each question,
the task is to rank all the candidate answers according to relevance. We
report precision@1 (P@1), mean average precision (mAP) and mean reciprocal rank
(MRR) in table~\ref{table:cqa-results}. Since there may be multiple paired answers
to a community posted question, the rank (for MRR) of a relevant answer is the
number of irrelevant answers ranked above it plus one. We observe that both 
SemQA-CQA and
SemQA-CQA-DS are able to significantly outperform other methods. This is 
expected since both of these methods were trained using CQA data and
hence is able to learn a good QA relevance function, whereas the 
% \sout{pre-trained sentence-transformers} 
sentence-transformers-STS-NLI were trained using STS and NLI tasks and they failed
to generalize. However, CQA pairs are significantly different from
the language of user queries and in the next section, we will evaluate
on those queries (the main goal of this paper).
\begin{table}[h]
\begin{tabular}{|ll|llll|}
\hline
                                         &   & \multicolumn{1}{c}{M0} & \multicolumn{1}{c}{M1} & \multicolumn{1}{c}{M2} & \multicolumn{1}{c|}{M3} \\ \hline
\multicolumn{1}{|l|}{\multirow{4}{*}{P@1}} & CA &      39.02                 &      52.87                 &      \textbf{74.10}             &  73.55   \\
\multicolumn{1}{|l|}{}                   & DE &        40.82               &        38.66               &        \textbf{73.04}             &   71.65 \\
\multicolumn{1}{|l|}{}                   & FR &        37.42               &        42.25               &        73.85             &   \textbf{75.34}\\
\multicolumn{1}{|l|}{}                   & IN &        26.51               &        35.67               &        53.05             &   \textbf{53.62}\\ \hline
\multicolumn{1}{|l|}{\multirow{4}{*}{mAP}} & CA &      41.02                 &      51.23                 &      \textbf{73.37}             &   72.46  \\
\multicolumn{1}{|l|}{}                   & DE &        45.24               &        43.35               &        \textbf{74.80}             &   73.04\\
\multicolumn{1}{|l|}{}                   & FR &        41.24               &        44.46               &        74.27             &   \textbf{75.38}\\
\multicolumn{1}{|l|}{}                   & IN &        43.93               &        51.12               &        \textbf{72.17}             &   71.89\\ \hline
\multicolumn{1}{|l|}{\multirow{4}{*}{MRR}} & CA &      31.21                 &      42.41                 &      \textbf{65.17}             &   64.17  \\
\multicolumn{1}{|l|}{}                   & DE &        37.05               &        35.48               &        \textbf{68.30}             &   66.20\\
\multicolumn{1}{|l|}{}                   & FR &        32.26               &        35.03               &        66.78             &   \textbf{67.37}\\
\multicolumn{1}{|l|}{}                   & IN &        34.70               &        41.23               &        58.44             &   \textbf{58.46}\\ \hline
\end{tabular}
\caption{Evaluation on CQA pairs. M0: BM25, M1: sentence-transformers-STS-NLI, M2: SemQA-CQA, M3: SemQA-CQA-DS.}
\label{table:cqa-results}
\end{table}

\begin{table*}[t]
\begin{tabular}{|l|l|l|}
\hline
\parbox[t]{3.5cm}{User query} & \parbox[t]{5.5cm}{Top CQA pair retrieved by SemQA-CQA-DS} & \parbox[t]{5.5cm}{Top CQA pair retrieved by E-commerce Baseline} \\ \hline
\multicolumn{3}{c}{Improving semantic relevance} \\ \hline
\parbox[t]{3.5cm}{Do you have size variation??? Like i need  this in bigger wood..} & 
    \parbox[t]{5.5cm}{\textbf{Q:} Is this available in still large size \\ \textbf{A:} yes .. available size is 7*5,8*6,9*7,12*9 in inches} & 
    \parbox[t]{5.5cm}{\textbf{Q:} Is the wood and print waterproof ? \\ \textbf{A:} YES} \\\hline
\parbox[t]{3.5cm}{It is compatible in gaming} & 
    \parbox[t]{5.5cm}{\textbf{Q:} Does it  run gta v \\ \textbf{A:} Yesss... Very fine} & 
    \parbox[t]{5.5cm}{\textbf{Q:} Is it compatible with Amd A6 processor ? \\ \textbf{A:} Yes it's compatible DDR4} \\\hline
\parbox[t]{3.5cm}{Total weight} & 
    \parbox[t]{5.5cm}{\textbf{Q:} Each 1 how to kgs \\ \textbf{A:} 10 kgs} & 
    \parbox[t]{5.5cm}{\textbf{Q:} Total diameter of the plates? \\ \textbf{A:} Plate Dia is 9.5 inches Hole Dia is 30 mm} \\ [0.75cm] \hline
\multicolumn{3}{c}{Improving coverage} \\ \hline 
\parbox[t]{3.5cm}{What is fabric} & 
    \parbox[t]{5.5cm}{\textbf{Q:} Which material is the scarf made up of \\ \textbf{A:} It is like soft satin silk} & 
    \parbox[t]{5.5cm}{No response} \\\hline 
\parbox[t]{3.5cm}{The dress with hands or seelveless} & 
    \parbox[t]{5.5cm}{\textbf{Q:} Is it sleeveless \\ \textbf{A:} we give a extra sleeves so u can attach or not..as ur wish} & 
    \parbox[t]{5.5cm}{No response} \\\hline 
\parbox[t]{3.5cm}{Bettry perfon} & 
    \parbox[t]{5.5cm}{\textbf{Q:} Batrrey capictiy \\ \textbf{A:} This Phone has a Wonderful 4000 Mah Battery with Battery Saver Options \& Can Watch videos continuously for 18 Hours!!!} & 
    \parbox[t]{5.5cm}{No response} \\ \hline
\end{tabular}
\caption{Qualitative examples.}
\label{table:qualitative-results}
\end{table*}

\noindent \textbf{Evaluation on User Queries:}
To evaluate on user queries, 
% \sout{we sample
% queries and top responses from the query logs of the instant QA system.}
we sample user queries (and their corresponding top responses) uniformly at random 
from the query logs of the instant QA system.
We also
retrieve the top responses generated by the different models we trained.
%The statistics of test data for each locale is described in table~\ref{}.
These query, response pairs are labeled as relevant or irrelevant
by a team of human annotators.
%A positive label implies the response is relevant to the query,
%and negative label means the response is irrelevant.
We use the area under
the precision recall curve (PR-AUC) as our quality metric. We report the
absolute percentage points change in PR-AUC with respect to the  
E-commerce 
Baseline in
table~\ref{table:lq-pr-auc-results} (+ve sign implies PR-AUC has improved and 
-ve sign implies PR-AUC
has decreased). We make the following observations: (1) the vanilla BM25
baseline performs the worst which is expected as it relies solely on syntactic
matches and fails to capture semantic intent; (2) both the 
% \sout{pre-trained sentence-
% transformers} 
sentence-transformers-STS-NLI and our SemQA-CQA models fail to generalize validating our
hypothesis that learning a general semantic matching model or a QA relevance
model is not sufficient to learn the nuances of user querying language; (3) 
the
SemQA-CQA-DS models significantly outperform all other models.% and improve
%the PR-AUC by upto xx absolute percentage points. 
There are two underlying
reasons for these improvements. Firstly, SemQA-CQA-DS is able to leverage the
semantic understanding capabilities (that Pretrained-Transformers and 
SemQA-CQA are also able to do), and secondly, SemQA-CQA-DS is also able to learn 
the nuances of the task
specific query language leading to a better relevance model between user 
queries and CQA pairs (that are
potential candidate responses).
%The language of the customer queries on the instant 
%answering system on product pages are quite different. They are usually short,
%ill-formed and often incomplete. Our distant supervision is able to learn this
%task specific query language and hence is able to learn a better and robust
%relevance model between customer queries and the community QA pairs (that are
%potential candidate responses).
\begin{table}[h]
\begin{tabular}{|l|llll|}
\hline
   & \multicolumn{1}{c}{M0} & \multicolumn{1}{c}{M1} & \multicolumn{1}{c}{M2} & \multicolumn{1}{c|}{M3} \\ \hline
CA & -19.75   & -1.11   & +1.53   & \textbf{+9.25}   \\
DE & -13.03   & -11.90   & +5.46   & \textbf{+12.15}   \\
FR & -11.66   & -4.54   & +4.93   & \textbf{+7.68}   \\
IN & -16.66   & -0.26   & +0.39   & \textbf{+4.37}   \\ \hline
\end{tabular}
\caption{PR-AUC on user queries evaluation set. M0: BM25, M1: sentence-transformers-STS-NLI, M2: SemQA-CQA, M3: SemQA-CQA-DS. Numbers denote the absolute
percentage points change with respect to the E-commerce Baseline.}
\label{table:lq-pr-auc-results}
\end{table}

Next, we do a qualitative analysis on the cases where SemQA-CQA-DS is able to
improve on the E-commerce Baseline. We identify two main areas of improvement:
(1) improving relevance in cases where the baseline fails to capture the
semantic intent, and
(2) improving coverage in cases where the baseline fails to retrieves any
response.
%(1)
%improving coverage by answering queries that the existing
%system can't answer as syntactic matches are not enough and (2) improving
%elevance by retrieving a semantically relevant answer in cases where the
%answer by the syntactic match based system is not relevant as it is not able to
%capture the semantic intent. 
We present examples of both cases in table~\ref{table:qualitative-results}. The
examples include cases where the language is ill-formed and incoherent and our
distantly supervised model still captures the intent and retrieve
relevant responses.

\noindent \subsection{Online Evaluation}
We also ran a large scale online A/B experiment with 50\% of the user traffic.
All locales were experimented at least for two weeks to ensure diversity in periodic patterns and have enough queries to achieve statistically 
significant conclusions (p-values < 0.01 in Chi-Square tests) about the improvement in metrics.
Here the SemQA-CQA-DS model is used to complement the existing E-commerce 
Baseline~\footnote{Details can't be disclosed due to proprietary information}
to improve the coverage of the system. There are two metrics of interest:
(1) the coverage (percentage of queries answered by the system), and (2) the
new question asking rate (percentage of queries for which even
after seeing the response, a user asks a question in the CQA
forum; if the relevance of the answers improves, the question asking rate should
decrease). We report the change in absolute percentage points with respect to
the E-commerce Baseline (for coverage +ve is better, and for question asking rate
-ve is better). The results are present in table~\ref{table:ab-test-results}. 
SemQA-CQA-DS was able
to improve coverage while reducing the rate of new questions posted by users
in all locales thereby showing the efficacy of our approach at scale.
\begin{table}[h]
\begin{tabular}{|l|cc|}
\hline
   & \multicolumn{1}{c}{Coverage} & \multicolumn{1}{c|}{\begin{tabular}[c]{@{}c@{}}Question Asking Rate\end{tabular}} \\ \hline
CA & +2.96   & -0.69      \\
DE & +3.12   & -0.44      \\
FR & +4.56   & -1.60      \\
IN & +6.92   & -0.97      \\ \hline
\end{tabular}
\caption{A/B test evaluation. Numbers denote the absolute
percentage points change of Treatment with respect to Control.}
\label{table:ab-test-results}
\end{table}
%SemQA-CQA-DS is able
%to improve the coverage by upto xx absolute percentage points, while
%simultaneously improving the question asking rate by xx absolute percentage 
%points.
% \begin{table}[h]
% \begin{tabular}{|l|cc|}
% \hline
%   & \multicolumn{1}{c}{Coverage} & \multicolumn{1}{c|}{\begin{tabular}[c]{@{}c@{}}Question Asking Rate\end{tabular}} \\ \hline
% CA & +2.96   & -0.69      \\
% DE & +3.12   & -0.44      \\
% FR & +4.56   & -1.60      \\
% IN & +6.92   & -0.97      \\ \hline
% \end{tabular}
% \caption{A/B test evaluation. Numbers denote the absolute
% percentage points change of Treatment with respect to Control.}
% \label{table:ab-test-results}
% \end{table}

%% file: conclusion.tex
\section{Conclusions \& Future Works}
In this paper we presented `SemQA', a practical transformer-based framework to
provide instant QA efficiently on the product pages of e-commerce
services. Given a user query, our framework directly retrieves the relevant 
CQA pairs from the product page, where user queries and CQA pairs
have significantly different language characteristics. Our model is able to 
learn a robust
relevance function between user queries and CQA pairs by learning 
representations that leverage the strengths of both syntactic and semantic
features, without the need for
any explicit human labeled data. Our model is able to scale to large scale
real-time e-commerce systems and at inference time only requires model encoding
of user queries for by index lookups, and 
candidate responses are encoded offline into the index in a space efficient
manner. Extensive offline evaluation shows our approach generalizes to multiple
locales spanning different languages with a PR-AUC gain by upto 12.15\% over
the existing system at a popular e-commerce service. We also ran a large scale
online A/B experiment with 50\% of the user traffic and our framework was able
to improve coverage by upto 6.92\% by complementing the existing system.

As a future direction, we would like to expand our SemQA system to include 
responses
from additional content on the product pages (reviews, descriptions etc.). We
believe some of the existing approaches to leverage reviews (discussed in 
section~\ref{sec:related}) can be used to complement our system to expand our
relevance model beyond CQA data. Another direction of research will be
to include features such as accuracy, sentiment, freshness etc. within our
proposed SemQA system's responses.

%% file: ack.tex
\section{Acknowledgements}
We thank all the anonymous reviewers for providing their valuable comments that helped us
improve the quality of our paper. We also thank our colleagues in the science, product, and engineering teams at Amazon for their valuable
inputs.